# IHUBERT: Vector-Based Semantic Deduplication and Domain-Balanced Pretraining for Persian Resources

Arash Ghafouri*, Mahdi Firouzmandi, Hossein Saberi and Mohammad Reza Hasani Ahangar

Department of Artificial Intelligence and Cognitive Science, Imam Hossein Comprehensive University, Tehran, Iran.

* Corresponding author E-mail: Ghafouri@ihu.ac.ir.

Contributing authors E-mail: Firouzmandi@ihu.ac.ir; Hsaberi@ihu.ac.ir; Mrhasani@ihu.ac.ir.

## Abstract

Persian pretrained language models (PLMs) are still limited by the scarcity of large-scale, high-quality pretraining corpora and by insufficient evaluation beyond standard classification and NER tasks. We present IHUBERT, a monolingual Persian PLM trained from scratch with the RoBERTa-base encoder (125M parameters) on a 45 GB curated subset of the Sepahr-Danesh collection (≈ 7–8B tokens). To improve corpus quality and reduce redundancy, we employ a multi-stage preprocessing pipeline that includes normalization, exact and near-duplicate removal, anonymization, and vector-database–based semantic deduplication for distribution balancing control across domains and registers. We additionally train a 139k-vocabulary BPE tokenizer on the full pretraining corpus to better capture Persian morphology and orthographic variation. IHUBERT is evaluated on seven Persian NLU benchmarks covering NER, sentiment analysis, topic classification, NLI, extractive question answering, and relation extraction, using task-standard metrics (entity-level F1, Macro-F1, EM/F1). IHUBERT achieves its strongest gains on extractive QA, ranking 1st on both PQuAD (F1 88.3542) and ParsiNLU-RC (F1 49.0987), and attains the best result on FarsTail (Macro-F1 0.8350). On NER and topic classification, it remains competitive (e.g., 0.8308 F1 on ParsTwiNER; 0.7953 Macro-F1 on DigiMag), while relation extraction remains the main remaining gap (0.6684 Macro-F1 on PERLEX). A controlled tokenizer ablation on the IHUBERT pretraining corpus shows that BPE yields slightly lower subword fragmentation than WordPiece at matched vocabulary size, supporting our tokenization design. Overall, IHUBERT advances Persian language modeling through semantically curated large-scale pretraining and broad evaluation across both classification and comprehension-oriented tasks.

## 1. Introduction

Pre-trained language models (PLMs) have become a central paradigm in natural language processing by enabling effective transfer learning across a wide range of downstream tasks. While high-resource languages have benefited substantially from large-scale corpora and continuous model development, Persian still faces notable limitations in both the scale and the curation quality of available pretraining data, as well as in comprehensive evaluation coverage. Multilingual PLMs such as mBERT [1] and XLM-R [2] provide baseline support for Persian, yet they often lag behind strong monolingual models due to shared capacity and vocabulary across many languages.

In response, the Persian NLP community has introduced several monolingual PLMs. ParsBERT[3] established an early strong baseline using a BERT-style architecture and a sizeable Persian corpus, showing clear gains over multilingual baselines on standard tasks. More recent efforts expanded data diversity and training recipes, including AriaBERT [4] (RoBERTa-style pretraining with BPE), FaBERT [5](large-scale blog pretraining with enhanced masking strategies), and models focusing on informal or domain-specific Persian such as FarSSiBERT [6] and SINA-BERT [7]. Despite this progress, two limitations frequently remain: (i) ensuring high-quality, non-redundant, and well-distributed pretraining data at scale, and (ii) evaluating models beyond classification/NER to include reasoning- and comprehension-oriented benchmarks (e.g., NLI and QA).

In this work, we introduce IHUBERT, a monolingual Persian PLM based on the RoBERTa-base transformer encoder and trained from scratch with a focus on data quality, tokenizer coverage, and broad evaluation. In particular, IHUBERT is pretrained on a 45GB curated subset (≈ 7–8B tokens) sampled from the larger Sepahr-Danesh collection. Instead of relying solely on surface-form filtering, we employ a vector-database (VDB)–based semantic deduplication and redundancy control pipeline to reduce near-duplicate content while preserving topical and stylistic diversity. In addition, we adopt a custom BPE tokenizer with a 139k-entry vocabulary, designed to reduce excessive subword fragmentation for Persian's morphology and orthographic variation.

Our contributions are summarized as follows:

- **Semantically curated large-scale Persian pretraining data:** We construct a 45GB (≈7–8B tokens) subset sampled from the Sepahr-Danesh corpus, applying rigorous quality controls including normalization, exact and near-duplicate removal, anonymization, and vector-database–based semantic deduplication to limit redundancy and improve distributional coverage. Figure 1 provides an overview of the proposed preprocessing and semantic filtering pipeline.

- **Morphology- and variation-aware tokenization:** We train a 139k-vocabulary BPE tokenizer on the same corpus to better capture Persian subword structure and reduce fragmentation relative to smaller-vocabulary setups.

- **RoBERTa-style pretraining from scratch:** IHUBERT follows the RoBERTa recipe (MLM-only, dynamic masking, no NSP) for robust representation learning [8].

- **Broad evaluation across Persian NLU:** We evaluate IHUBERT on seven Persian benchmarks covering token-level sequence labeling (NER), text classification (sentiment, topic classification, and NLI), extractive QA, and relation extraction, using task-standard metrics (entity-level F1, Macro-F1, and EM/F1).

The remainder of this paper is organized as follows. Section 2 reviews related work on Persian pretrained models and corpus construction. Section 3 describes our methodology, including the construction of the 45 GB Sepahr-Danesh subset and the multi-stage preprocessing pipeline (cleaning, exact and near-duplicate removal, vector-database semantic deduplication, TF-IDF prioritization, and tokenizer training); Figure 1 illustrates this pipeline. Section 4 details the IHUBERT design: the 139k-entry BPE tokenizer, RoBERTa-base architecture, and the pretraining setup (hardware, optimization, and convergence diagnostics), with an overview diagram in Figure 2. Section 5 presents the evaluation protocol, datasets, metrics, and baselines. Section 6 reports experimental results across NER, classification, extractive QA, and relation extraction, followed by task-level discussion. Section 7 gives ablations and additional analyses (notably the controlled BPE vs WordPiece tokenizer study). Finally, Section 8 concludes and outlines directions for future work.

## 2. Related Work

**Persian Pre-trained Language Models**. Early Persian PLMs largely followed the BERT [1] architecture. ParsBERT [3], trained on approximately 14GB of web and informal text, demonstrated strong performance on sentiment classification, news categorization, and NER, achieving up to 93–98% F1 on PEYMA and Arman. Subsequent models expanded domain and style coverage. AriaBERT [4] adopted the RoBERTa[8] architecture and BPE tokenization, leveraging a 32GB corpus of formal, conversational, medical, poetic, and user-generated text, and reported consistent improvements over ParsBERT [3] and multilingual baselines. Models trained on large-scale social media corpora—such as FarSSiBERT[6] (104M tweets) and FaBERT[5] (6.8B tokens from HmBlogs)—showed that domain-specific pretraining substantially enhances performance on colloquial and noisy text. Domain-adapted variants such as SINA-BERT[7] further demonstrated that targeted medical corpora improve specialized QA and analysis tasks. More recent efforts (e.g., TookaBERT [9] and HakimBERT[10]) incorporate modern training techniques, larger vocabularies, and broader corpora, yielding competitive or superior results across NLU benchmarks. Despite this progress, limitations remain regarding corpus diversity, tokenizer coverage, and evaluation breadth.

***Table 1***. Summary of major Persian monolingual PLMs.

| ***Model*** | ***Architecture / Params*** | ***Tokenizer (type, vocab)*** | ***Pretraining corpus (size & domain)*** | ***Primary focus / notes*** |
|---|---|---|---|---|
| ParsBERT [3] | BERT-base (110M) | **WordPiece** (100k) | 14GB web+wiki | strong SA/NER, > mBERT. |
| AriaBERT [4] | RoBERTa-base (125M) | **BPE** (60k) | 32GB mixed (formal+informal ) | Improved SA/TC/stance vs ParsBERT/XLM-R. |
| FarSSiBERT [6] | BERT-base | Informal subword | 48GB (104M tweets) | Social-media model; superior informal similarity. |
| FaBERT [5] | BERT-base (124M) | **WordPiece** (50k) | 6.8B-token HmBlogs | Strong on 12 NLU tasks (NLI, QA, NER). |
| SINA-BERT [7] | BERT-base (110M) | ParsBERT WP | 2.8M medical docs | Domain medical PLM; better medical QA. |
| TookaBERT [9] | Base/Large PLMs | **BPE** (48k) | Large web mix | Large model; +2–3 pts over prior PLMs. |
| HakimBERT [10] | BERT/RetroMAE encoders | **WordPiece** (50k) | 11B-token Corpesia + pairs | SOTA Persian embeddings; +8.5% FaMTEB. |

| IHUBERT (ours) | RoBERTa-base (125M) | BPE (139k) | 45GB (7–8B tokens) | Full-scratch PLM; broad NLU+reasoning gains. |
|---|---|---|---|---|

**Multilingual Models for Persian.** Multilingual PLMs such as mBERT [1] and XLM-R[2] provide baseline capabilities for Persian; however, extensive empirical evidence indicates that monolingual training generally outperforms multilingual counterparts due to language-specific morphology and vocabulary. Prior work consistently reports gaps between mBERT/XLM-R and specialized Persian models, a trend confirmed in our experiments where IHUBERT and ParsBERT [3] outperform multilingual baselines across tasks.

**Monolingual PLMs in Other Languages.** The development of IHUBERT aligns with successful monolingual RoBERTa- or BERT-based models in other languages, including AraBERT[11] (Arabic), HerBERT[12] (Polish), BERTimbau[13] (Portuguese), Kr-BERT[14] (Korean), and GottBERT[15] (German). These works demonstrate that language-specific corpora, morphology-aware tokenization, and pretraining from scratch substantially improve downstream accuracy. IHUBERT follows this paradigm, employing the RoBERTa training recipe, dynamic masking, and a large BPE vocabulary designed for Persian's rich morphology.

**Persian Benchmarks and Datasets.** A diverse set of benchmarks has emerged for Persian NLU. Classification datasets include Digikala Magazine and Persian News; offensive language detection is supported by Pars-OFF [16]; sentiment and emotion analysis utilize DeepSentiPers, Digikala, SnappFood, Taghche, and Pars-ABSA [17]; irony detection is enabled by MirasIrony; and stance detection datasets such as FarExStance [18] are available. Higher-level reasoning and semantic understanding tasks are supported by FarsTail [19] for natural language inference (NLI), FarSick [20] for semantic textual similarity (STS), and PQuAD [21] / PCoQA [22] for extractive and conversational machine reading comprehension.

Prior models primarily focused on classification and NER, whereas comprehensive evaluation on reasoning-oriented tasks (e.g., NLI, QA, STS) remains limited. IHUBERT expands evaluation beyond standard classification/NER by additionally covering NLI, extractive QA, and relation extraction, alongside core classification and NER benchmarks.

***Table 2.*** Key Persian NLU datasets used in prior work and in IHUBERT evaluation.

| ***Dataset*** | ***Task*** | ***Used by*** |
|---|---|---|
| ParsTwiNER [23] | NER | FaBERT, TookaBERT, HakimBERT, **IHUBERT** |
| DeepSentiPers [24] | Sentiment | ParsBERT, FaBERT, HakimBERT, TookaBERT, **IHUBERT** |
| Digikala Magazine | Topic Classification | ParsBERT, AriaBERT, **IHUBERT** |
| FarsTail [19] | NLI | FaBERT, TookaBERT, HakimBERT, **IHUBERT** |
| ParsiNLU [25] | Paraphrase / Duplicate | FaBERT, TookaBERT, HakimBERT, **IHUBERT** |
| PQuAD [21] | QA (Extractive) | FaBERT, TookaBERT, HakimBERT, **IHUBERT** |
| PERLEX [26] | Relation Extraction (RE) | **IHUBERT** |

**Research Gaps and Positioning of IHUBERT.** Existing Persian PLMs generally lack (i) a large, semantically filtered corpus combining formal, conversational, and domain-specific text; (ii) a high-capacity 139k-token BPE vocabulary; (iii) a vector-database–driven semantic deduplication pipeline; and (iv) full pretraining from scratch on 7–8B tokens. IHUBERT addresses these gaps by integrating a diverse 45GB corpus, advanced preprocessing, and broad evaluation across both classification and reasoning tasks, thereby advancing the state of Persian language modeling.

# 3. Methodology/ Dataset & Preprocessing

## 3.1. Dataset Description

### 3.1.1. Overview

IHUBERT was pretrained on a curated 45GB subset of the Sepahr-Danesh corpus. The full corpus contains over 55B tokens across more than 100M documents, but a controlled subset of 7–8B tokens was selected to ensure data quality, reduce redundancy, and maintain a balanced distribution across domains. This scale provides sufficient linguistic coverage for training a RoBERTa-base model while preserving computational efficiency.

### 3.1.2. Data Composition

The final dataset integrates three major categories:

**Formal Texts.** Newswire articles from 50 agencies, Persian Wikipedia, encyclopedic and academic documents, and selected literary texts. These sources provide well-structured grammar and support stable syntactic and lexical modeling.

**Informal Texts.** Social-media content (Twitter, Telegram), user reviews, and conversational text containing slang, typos, and emoji usage. These sources enable the model to generalize to spoken-style and noisy user-generated text.

**Mixed and Domain-Specific Texts.** Medical, legal, governmental, technical, and semi-formal web content. These materials broaden lexical coverage and ensure domain adaptability.

### 3.1.3. Sampling Strategy

Subset selection followed a controlled sampling procedure designed to:

– preserve stylistic and topical diversity across more than 1,000 sources.

– eliminate low-quality, machine-generated, incomplete, or duplicated documents.

– remove semantic redundancy via vector-based similarity filtering.

– maintain paragraph-level continuity consistent with the construction guidelines of Sepahr-Danesh.

This process ensures a dataset that is representative, high-quality, and suitable for large-scale masked-language modeling.

#### 3.1.4. Final Corpus Distribution

The resulting composition is approximately:

– 40% formal text,

– 40% informal and conversational text,

– 20% mixed and domain-specific text.

This balanced distribution supports the design of the 139k-entry BPE tokenizer and enables IHUBERT to perform robustly across both formal and informal registers.

#### 3.1.5. Comparative Advantages

Compared with existing Persian corpora—such as the ParsBERT[3] corpus (~14GB) and AriaBERT [4] corpus (~32GB)—this dataset provides:

– substantially broader domain and register coverage,

– preserved paragraph structure instead of fragmented web text,

– simultaneous inclusion of formal, colloquial, and technical language,

– cleaner metadata and well-defined document boundaries.

These characteristics enhance representativeness, support stable convergence, and enable improved downstream performance.

### 3.2. Preprocessing

#### 3.2.1. Overview

A multi-stage preprocessing pipeline was designed to ensure that the input to IHUBERT exhibits adequate linguistic quality, structural consistency, and balanced statistical distribution. The pipeline was applied to a 45GB subset of the Sepahr-Danesh corpus, corresponding to approximately 7–8 billion tokens drawn from diverse formal and informal Persian sources. The preprocessing workflow integrates text cleaning, normalization, semantic filtering, deduplication, TF-IDF scoring, and preparation of data for BPE tokenizer training and masked-language modeling.

#### 3.2.2. Data Sources

Two primary data families were incorporated:

1. High-quality sources, including Wikipedia, books, scientific articles, and curated newswire text.
2. General-domain sources, such as social-media posts, weblogs, comments, and informal user-generated content.

Both streams enter the preprocessing pipeline, though the depth of cleaning differs due to variation in noise levels and structural quality.

#### 3.2.3. Basic Cleaning

All documents underwent a baseline cleaning phase that removed HTML tags, scripts, boilerplate code, advertisements, and malformed markup. Additional steps included spelling and punctuation normalization, correction of encoding inconsistencies, removal of non-Persian or extremely short fragments, and exact-duplicate filtering at the document and sentence levels. These operations ensured that all retained text met a minimum standard of linguistic coherence.

#### 3.2.4. Semantic Cleaning and Distribution Control

A key component of the preprocessing pipeline is a semantic deduplication module built on a vector-database retrieval system. A sample of high-quality cleaned text was embedded using a pretrained encoder and indexed as a semantic reference set. Each remaining candidate document was vectorized and compared against this set using similarity search. Documents exceeding a similarity threshold were discarded to prevent semantic redundancy, domain imbalance (e.g., replicated news stories), and inadvertent memorization. This process served to normalize the final corpus distribution across topics and styles.

#### 3.2.5. Hard Cleaning and Main Dataset Construction

After semantic filtering, a stricter cleaning stage was applied. This included the removal of long repetitive templates (such as website footers or boilerplate paragraphs), exclusion of low-quality or syntactically degraded text, and enforcement of constraints on document length and noise ratio. The output of this stage constitutes the Main Training Dataset, which stabilizes at approximately 45GB (≈7–8B tokens).

#### 3.2.6. Custom BPE Tokenizer

Once the cleaned dataset was finalized, a custom Byte-Pair Encoding tokenizer was trained. Multiple candidate vocabularies (e.g., 48k and 120k) were initially explored, after which the final tokenizer was trained on the entire 45GB corpus, producing a 139k-token vocabulary. This vocabulary size provides an effective compromise between reducing over-segmentation in morphologically rich Persian text and maintaining adequate generalization across formal and informal domains.

#### 3.2.7. TF-IDF Scoring and Content Prioritization

To emphasize linguistically informative samples, documents were tokenized and transformed into TF-IDF space. Sentences with extremely low mean TF-IDF scores—dominated by high-frequency functional tokens—were down-weighted or removed, while semantically rich texts were retained. Combined with semantic deduplication, this ensured that the final corpus preserved both breadth and informational depth.

#### 3.2.8. Summary

The final preprocessing pipeline integrates (i) multi-source data ingestion, (ii) structural and linguistic cleaning, (iii) semantic deduplication via vector-database retrieval, (iv) hard cleaning and domain balancing, (v) TF-IDF-based content filtering, and (vi) the construction of a 139k-vocabulary BPE tokenizer. The resulting 45GB corpus is a high-quality, well-balanced dataset suitable for large-scale masked-language-model pretraining.

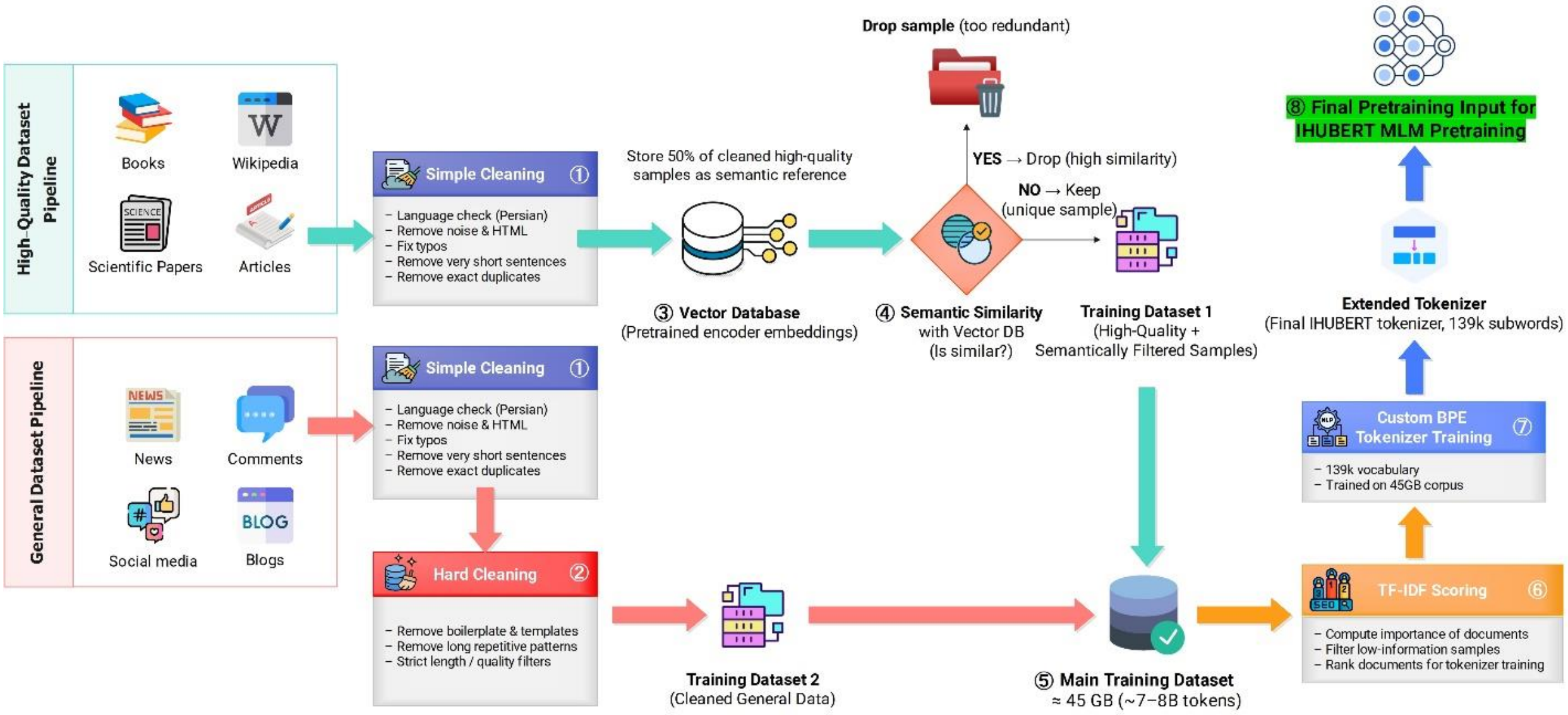

***Figure 1.*** The IHUBERT preprocessing pipeline. High-quality and general corpora undergo simple cleaning, with the general corpus additionally processed through hard-cleaning. Half of the cleaned high-quality samples are stored in a vector database to enable semantic similarity filtering, producing a refined Training Dataset 1. The cleaned general corpus forms Training Dataset 2. Both datasets are merged into the final ~45 GB corpus (~7–8B tokens). TF-IDF scoring selects informative samples for tokenizer construction, and a 139k-subword BPE tokenizer is trained on the corpus. The resulting dataset and tokenizer serve as the final input for IHUBERT's MLM pretraining.

***Table 3***. Comparison of preprocessing strategies (prior Persian PLMs vs. IHUBERT)

| *Model* | *Text norm. & cleaning* | *Lang./script filter* | *Deduplication* | *Privacy / semantic filtering* |
|---|---|---|---|---|
| ParsBERT [3] | ✓ | ✓ | — | — |
| AriaBERT [4] | △ | — | — | — |
| FarSSiBERT [6] | ✓ | ✓ | △ | — |
| FaBERT [5] | ✓ | ✓ | △ | — |
| SINA-BERT [7] | ✓ | ✓ | — | — |
| TookaBERT [9] | ✓ | ✓ | — | — |
| HakimBERT [10] | ✓ | ✓ | △ | — |
| **IHUBERT (ours)** | ✓ | ✓ | **✓ (exact + near)** | **✓ (PII + VDB semantic)** |

**Legend:** ✓ = explicitly reported, △ = partially described, — = not reported.

# 4. Tokenizer and Architecture (IHUBERT design)

## 4.1. Byte-Pair Encoding (BPE) rationale

Byte-Pair Encoding (BPE) is a frequency-based subword segmentation algorithm that incrementally merges the most common symbol pairs to form a compact and expressive vocabulary. By decomposing rare or unseen words into familiar subword units, BPE effectively mitigates the out-of-vocabulary (OOV) problem while preserving meaningful morphological cues.

This approach enables a favorable balance between vocabulary size and computational efficiency, which is essential for large-scale pretraining. The method is particularly well-suited for morphologically rich languages such as Persian, where clitic attachment, affix stacking, and inconsistent spacing produce extensive surface-form variation. Given the noisy and domain-diverse nature of our large corpus, BPE provides a stable and robust tokenization strategy that allows the model to generalize across formal, informal, and user-generated text.

For IHUBERT, we trained the tokenizer from scratch on the full pretraining corpus to obtain a domain-comprehensive and linguistically coherent subword vocabulary. Using the BPE algorithm, we constructed a 139k-entry vocabulary designed to capture the rich inflectional patterns, clitic combinations, and orthographic variation characteristic of modern Persian. The enlarged vocabulary enables more accurate segmentation of high-frequency stems while limiting excessive fragmentation in informal or user-generated content. All preprocessing steps—including normalization, script unification, and masking of sensitive patterns—were applied prior to tokenizer training to ensure consistent subword statistics. This configuration provides IHUBERT with broad lexical coverage and stable tokenization across heterogeneous domains, supporting robust downstream NLU performance. In Section 7.1 , we further analyze this design choice via a controlled BPE vs WordPiece tokenizer ablation on the IHUBERT pretraining corpus, showing that BPE yields slightly lower subword fragmentation under a matched vocabulary size.

In addition to the structural advantages of BPE, it is worth noting that while BERT uses WordPiece, many newer Transformer LMs commonly adopt BPE [9]; we follow this practice for Persian and additionally confirm via our tokenizer ablation that BPE yields slightly lower fragmentation than WordPiece at matched vocabulary size. The IHUBERT tokenizer is uncased—consistent with the characteristics of the Persian script—and is designed to cover Persian letters, digits, punctuation, and frequently occurring emojis and symbols commonly found in social media text. Compared to the ~32k WordPiece vocabulary used by ParsBERT [3], the 139k-entry BPE vocabulary learned for IHUBERT offers substantially broader lexical coverage and reduces oversplitting in both formal and user-generated content. The tokenizer also produces linguistically meaningful segmentations; for example, the word "کتابهایمان" ("our books") is decomposed into subword units that separately capture the stem "کتاب" and its attached affixes. These properties collectively contribute to a more expressive and domain-robust tokenization scheme that benefits downstream NLU tasks.

### 4.2. IHUBERT (RoBERTa-base) Architecture Details

IHUBERT adopts the RoBERTa-base architecture [8], a robustly optimized variant of the original BERT-base Transformer encoder. The model consists of 12 Transformer encoder layers, each containing 12 self-attention heads, a hidden size of 768, and a 3072-dimensional feed-forward network, resulting in approximately 125 million parameters, comparable to BERT-base. IHUBERT's parameters were randomly initialized and the model was trained from scratch on our Persian pretraining corpus, rather than continuing from any multilingual checkpoint.

The choice of RoBERTa over BERT is motivated by its improved training strategy—removing Next Sentence Prediction (NSP), relying exclusively on Masked Language Modeling (MLM), adopting dynamic masking, and benefiting from larger batch sizes and longer training schedules, all of which have been shown to improve downstream robustness when sufficient data is available[8]. Consistent with the RoBERTa pretraining objective, 15% of input tokens are selected for masking, of which 80% are replaced with [MASK], 10% with random tokens, and 10% left unchanged. The model then predicts the original identity of these perturbed tokens, encouraging IHUBERT to learn rich bidirectional contextual representations.

IHUBERT follows the same architectural optimizations introduced in RoBERTa, including the GELU activation function, Layer Normalization, and dropout (0.1) applied across layers to ensure stable training. The maximum input length during pretraining is set to 512 subword tokens, enabling the model to process multi-sentence or paragraph-level documents without truncation. Positional embeddings of length 514 are used, consistent with RoBERTa's extended token indexing.

In summary, IHUBERT maintains the same architecture and parameterization as RoBERTa-base, while benefiting from a Persian-specific tokenizer with a 139k-entry BPE vocabulary tailored to the morphological and orthographic characteristics of Persian. Our tokenizer ablation (Section 7.1) compares BPE and WordPiece under matched conditions on the IHUBERT pretraining corpus and shows that BPE yields slightly lower subword fragmentation (e.g., fewer subwords per word and a lower fraction of words split into multiple pieces) at the same vocabulary size. This behavior aligns with the goal of reducing excessive segmentation for morphologically rich Persian, consistent with observations reported for similar languages.

The following section details the training setup and optimization strategy used for IHUBERT.

## 4.3. Training Setup

### 4.3.1. Training Framework and Data

IHUBERT was pretrained using a dedicated large-scale pipeline built on the HuggingFace Transformers framework. Pretraining was performed on a 45GB sampled subset of the Sepahr-Danesh corpus, containing approximately 7–8 billion tokens drawn from diverse formal, conversational, and user-generated domains. This subset provides sufficient linguistic coverage for robust model initialization while balancing compute constraints.

### 4.3.2. Tokenizer

We employ a custom BPE tokenizer trained on the full 45GB corpus, yielding a 139k-entry vocabulary tailored to Persian morphology. The larger vocabulary helps reduce excessive subword fragmentation in morphologically rich settings and aligns with the linguistic properties characterized in the Sepahr-Danesh dataset.

### 4.3.3. Hardware and Distributed Training

Pretraining was carried out on a two-GPU configuration consisting of one NVIDIA A100 80GB and one NVIDIA A100 40GB GPU. We used Distributed Data Parallel (DDP) together with mixed-precision (FP16) training to efficiently utilize the asymmetric hardware setup.

### 4.3.4. Optimization Strategy

IHUBERT was pretrained for 2 epochs on the curated 45GB corpus (≈7–8B tokens). Following the RoBERTa pretraining recipe [8], optimization was performed using:

- **Optimizer:** AdamW ($\beta_1 = 0.9$, $\beta_2 = 0.98$)

- **Weight decay:** 0.01
- **Peak learning rate:** $1\times10^{-4}$
- **Warmup:** first 10,000 steps (linear warmup)
- **LR schedule:** linear decay
- **Precision:** mixed precision (FP16)
- **Batching:** gradient accumulation to reach an effective global batch size of ~1,000–1,200 sequences
- **Max sequence length:** 512
- **Dropout:** 0.1
- **Positional embeddings:** 514 (RoBERTa-style)

#### 4.3.5. Training Time and Throughput

On our two-GPU A100 setup (80GB + 40GB) with DDP and FP16, pretraining took approximately 6–7 days per epoch (12–14 days total), corresponding to an observed throughput of roughly ~11–15k tokens/s under our effective batch size and sequence length.

#### 4.3.6. Monitoring and Convergence

Training loss was monitored continuously, and validation MLM loss was evaluated periodically on a 10-million-token held-out split drawn from the same sources. Validation loss began to plateau after ~1.5 epochs, indicating stable convergence with no signs of overfitting. The final checkpoint after the second epoch was selected for all downstream experiments.

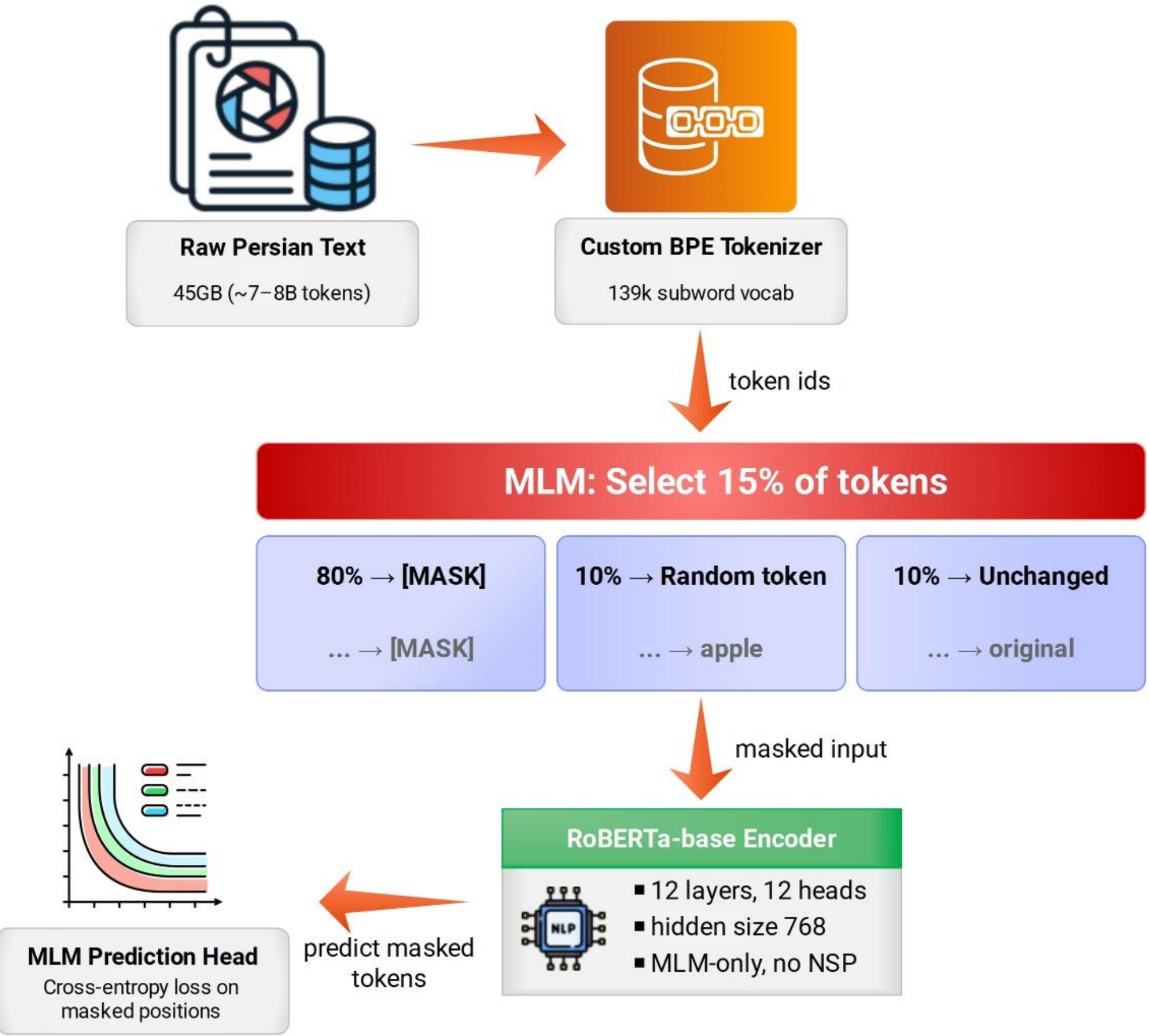

***Figure 2.*** IHUBERT pretraining overview: raw Persian text is tokenized using a 139k-vocabulary BPE tokenizer, dynamically masked following the RoBERTa MLM strategy (15% token selection; 80/10/10 replacement), and processed by a RoBERTa-base encoder trained from scratch with the MLM objective.

# 5. Evaluation Protocol

To assess IHUBERT, we conduct a broad evaluation across standard Persian NLU tasks. For each task, IHUBERT is fine-tuned on the task-specific training split and evaluated on the corresponding test split. This section specifies the downstream tasks and datasets, evaluation metrics, baseline models, and the fine-tuning protocol used to ensure fair comparisons.

## 5.1. Downstream Tasks and Datasets

Our evaluation covers several core Persian NLU tasks, each instantiated by a standard benchmark dataset:

(1) Named Entity Recognition (NER) on ParsTwiNER [23], formulated as token-level sequence labeling to identify entity spans and types in Persian social-media text.

(2) Sentiment Analysis on DeepSentiPers [24], a widely used sentiment benchmark for sentence/document-level polarity classification.

(3) Topic Classification on Digikala Magazine, a multi-class document classification task over magazine articles spanning diverse topical categories.

(4) Natural Language Inference (NLI) on FarsTail [19], formulated as a 3-way classification problem (entailment/neutral/contradiction) over premise–hypothesis pairs.

(5) Reading Comprehension (RC) on ParsiNLU [25] (ParsiNLU-RC), an extractive QA task in which the model predicts an answer span (or no-answer) given a question and a context passage.

(6) Extractive Question Answering (MRC) on PQuAD [21], where the model predicts an answer span (or no-answer) given a question and a context passage.

(7) Relation Extraction (RE) on PERLEX [26], predicting the relation type between entity mentions in context (Persian–English bilingual gold dataset).

## 5.2. Metrics

We follow standard evaluation protocols for each benchmark. For NER on ParsTwiNER [23], we report entity-level Precision, Recall, and F1 using span-based CoNLL-style evaluation. For classification tasks—sentiment on DeepSentiPers [24], topic classification on Digikala Magazine, NLI on FarsTail [19], and paraphrase/duplicate detection on ParsiNLU [25]—we report Accuracy and Macro-F1 (macro-averaged over labels), where Macro-F1 is emphasized to provide a class-balanced measure under label skew. For extractive QA on PQuAD [21], we report the standard Exact Match (EM) and token-level F1 scores (SQuAD-style). For relation extraction on PERLEX [26], we report Macro-F1 over relation labels (with Accuracy reported as an auxiliary metric), consistent with multi-class relation classification practice.

## 5.3. Baselines

To contextualize IHUBERT's performance, we compare against a set of strong Persian and multilingual pretrained encoders that are widely used as reference points in Persian NLU. For Persian monolingual baselines, we include ParsBERT [3] and AriaBERT [4] as established general-purpose Persian PLMs, and additionally consider more recent Persian checkpoints, including FaBERT [5] (blog-pretraining with mixed formality) and TookaBERT [9] as a recent Persian NLU-oriented BERT family. We also include Shiraz (lifeweb-ai/shiraz) as a publicly released Persian language model trained on large-scale curated data. As multilingual baselines, we report results for mBERT [1] and XLM-R [2], which provide strong cross-lingual reference performance and are commonly used in Persian benchmark comparisons. Unless otherwise stated, all baselines are used as publicly available checkpoints and are fine-tuned under the same downstream protocol as IHUBERT to ensure a fair comparison.

## 5.4. Fine-tuning Protocol & Reproducibility

To ensure a fair comparison, we fine-tune all baselines and IHUBERT using a unified training pipeline and consistent evaluation scripts per task. For each benchmark, we use the dataset's predefined splits when available; otherwise we create fixed splits with a constant random seed. Unless stated otherwise, we fine-tune models for 3 epochs with weight decay = 0.01 and seed = 42, changing only the underlying pretrained checkpoint while keeping the downstream task formulation identical across models.

### 5.5. Task Formulations

For sequence classification tasks (DeepSentiPers [24] sentiment; Digikala Magazine topic classification; FarsTail [19] NLI), we use a standard sequence classification head. For token classification (ParsTwiNER [23] NER), we use a token classification head with BIO tagging. For extractive question answering (PQuAD [21]; and the ParsiNLU [25] Reading Comprehension subset), we use span prediction (start/end) with SQuAD-style preprocessing; PQuAD [21] is handled in the SQuAD v2 setting with unanswerable questions. For relation extraction (PERLEX [26]), we train a multi-class relation classifier and report metrics on the official test set.

### 5.6. Dataset-specific Training Configurations

I. ParsTwiNER [23] (NER): max sequence length 128, learning rate 2e-5, batch size 16 (train) / 32 (eval), warmup ratio 0.06, evaluation and checkpointing per epoch, and best checkpoint selection based on overall entity-level F1; mixed precision (FP16) is enabled when supported.

II. DeepSentiPers [24] (5-class sentiment): stratified split 80/10/10 (train/validation/test), max length 128, learning rate 2e-5, batch size 16 (train) / 32 (eval), warmup ratio 0.06, best checkpoint selected by Macro-F1, FP16 when supported.

III. Digikala Magazine (topic classification): stratified split 70/15/15, max length 256, learning rate 2e-5, batch size 16, dynamic padding, and evaluation on the fixed test split; seed 42.

IV. FarsTail [19] (NLI): official split sizes (Train/Validation/Test = 7266/1537/1564), max length 256, learning rate 2e-5, batch size 16, evaluation each epoch; (no model checkpoint saving in this run).

V. ParsiNLU [25] Reading Comprehension (extractive QA): SQuAD-style span QA with sliding window preprocessing (max length 384, doc stride 128), learning rate 3e-5, batch size 16 (train) / 16 (eval), n-best 20, and max answer length 30; evaluation per epoch (no checkpoint saving).

VI. PQuAD [21] (extractive QA, SQuAD v2): max length 384, doc stride 128, learning rate 3e-5, effective batch size ≈ 16 via per-device batch 4 and gradient accumulation 4, n-best 20, max answer length 30, and null-answer threshold 0.0; FP16 enabled on GPU.

VII. PERLEX [26] (relation extraction): entity markers are converted into special tokens [E1] [/E1] [E2] [/E2] added to the tokenizer; max length 128, learning rate 2e-5, batch size 16, 10% validation split from training (with rare-label handling to avoid stratification failure), and evaluation on the fixed test set; (no checkpoint saving).

Overall, this protocol isolates the effect of the pretrained encoder by keeping downstream heads, preprocessing, and training settings consistent within each task, thereby enabling reproducible and fair cross-model comparisons.

## 6. Experimental Results & Analysis

We report results on seven Persian NLU benchmarks spanning token-level labeling, text classification, extractive QA, and relation extraction. For each task, we use the metrics defined in Section Metrics and rank models primarily by entity-level F1 (NER), Macro-F1 (classification and RE), and F1 (QA).

## 6.1. Overall Comparison

Across the seven benchmarks, IHUBERT achieves its strongest gains on extractive question answering, ranking 1st on both PQuAD [21] and ParsiNLU-RC [25] by F1. On NER and classification, IHUBERT remains competitive and typically ranks among the top group, while relation extraction (PERLEX [26]) is comparatively more challenging for IHUBERT in this evaluation and is led by TookaBERT-base.

## 6.2. Named Entity Recognition (ParsTwiNER)

Table 4 reports entity-level Precision/Recall/F1 (CoNLL-style, span-based). TookaBERT-base [9] achieves the best F1 (0.8480), while IHUBERT is second with 0.8308 F1, indicating strong sequence labeling performance and robust span detection on social-style Persian text.

***Table 4***. ParsTwiNER [23] (NER) entity-level results.

| ***Model*** | ***Precision*** | ***Recall*** | ***F1*** | ***Accuracy*** |
|---|---|---|---|---|
| **IHUBERT (ours)** | **0.8086** | **0.8543** | **0.8308** | **0.9751** |
| TookaBERT-base [9] | 0.8324 | 0.8642 | 0.8480 | 0.9765 |
| ParsBERT [3] | 0.7966 | 0.8162 | 0.8063 | 0.9714 |
| AriaBERT [4] | 0.6616 | 0.6974 | 0.6790 | 0.9569 |
| FaBERT [5] | 0.7760 | 0.8242 | 0.7994 | 0.9716 |
| Shiraz (lifeweb-ai/shiraz) | 0.7995 | 0.8408 | 0.8197 | 0.9745 |
| mBERT [1] | 0.7338 | 0.7618 | 0.7476 | 0.9626 |
| XLM-RoBERTa [2] | 0.7813 | 0.8481 | 0.8133 | 0.9724 |

## 6.3. Text Classification (DeepSentiPers, DigiMag, FarsTail)

Table 5 summarizes results for sentiment (DeepSentiPers [24], 5-class), topic classification (DigiMag, 7-class), and NLI (FarsTail [19], 3-class). We emphasize Macro-F1 for class-balanced comparison.

- DeepSentiPers [24]: XLM-RoBERTa [2] ranks first by Macro-F1 (0.5738), with IHUBERT close behind (0.5602).
- DigiMag: FaBERT [5] is best by Macro-F1 (0.7981), while IHUBERT is a near-tie (0.7953).
- FarsTail [19]: IHUBERT achieves the best Macro-F1 (0.8350) and Accuracy (0.8345), slightly outperforming TookaBERT-base [9] (Macro-F1 0.8322).

***Table 5***. Classification results (Accuracy / Macro-F1)

| *Model* | *DeepSentiPers [24]* | | *DigiMag* | | *FarsTail [19]* | |
|---|---|---|---|---|---|---|
| | *Acc* | *Macro-F1* | *Acc* | *Macro-F1* | *Acc* | *Macro-F1* |
| **IHUBERT (ours)** | **0.7127** | **0.5602** | **0.9444** | **0.7953** | **0.8345** | **0.8350** |
| TookaBERT-base [9] | 0.7253 | 0.5733 | 0.9515 | 0.7920 | 0.8331 | 0.8322 |
| ParsBERT [3] | 0.6804 | 0.5290 | 0.9484 | 0.7897 | 0.8286 | 0.8270 |
| AriaBERT [4] | 0.6912 | 0.5370 | 0.9444 | 0.7881 | 0.7141 | 0.7142 |
| FaBERT [5] | 0.7324 | 0.5685 | 0.9523 | 0.7981 | 0.8312 | 0.8302 |
| Shiraz (lifeweb-ai/shiraz) | 0.6822 | 0.5064 | 0.9460 | 0.7943 | 0.7001 | 0.7020 |
| mBERT [1] | 0.6391 | 0.4846 | 0.9374 | 0.7771 | 0.8312 | 0.8300 |
| XLM-RoBERTa [2] | 0.7342 | 0.5738 | 0.9444 | 0.7871 | 0.8159 | 0.8156 |

**Discussion.** The classification benchmarks show that performance is strongly benchmark-dependent: multilingual pretraining (XLM-RoBERTa [2]) is particularly effective for sentiment, while Persian-specific pretraining variants (FaBERT [5], TookaBERT-base [9]) lead on topic/NLI. IHUBERT stays consistently close to the leading models, suggesting strong generalization across diverse classification regimes.

## 6.4. Extractive Question Answering (ParsiNLU-RC [25], PQuAD)

Table 6 reports extractive QA results. IHUBERT ranks first on both QA benchmarks, with the best F1 on ParsiNLU-RC [25] (49.0987) and PQuAD [21] (88.3542).

***Table 6***. Extractive QA results (EM / F1)

| *Model* | ***ParsiNLU-RC* [25]** | | *PQuAD* | |
|---|---|---|---|---|
| | *EM* | *F1* | *EM* | *F1* |
| **IHUBERT (ours)** | **27.3684** | **49.0987** | **75.9810** | **88.3542** |
| TookaBERT-base [9] | 22.1053 | 43.0486 | 75.5311 | 87.7445 |
| ParsBERT [3] | 20.8772 | 39.1235 | 71.6321 | 85.4047 |
| AriaBERT [4] | 14.0351 | 33.9124 | 69.6076 | 82.7194 |
| FaBERT [5] | 0.8772 | 5.0492 | 28.5304 | 39.6162 |
| Shiraz (lifeweb-ai/shiraz) | 12.9825 | 33.0907 | 68.4954 | 83.8890 |
| mBERT [1] | 23.1579 | 42.5636 | 73.2692 | 86.1829 |
| XLM-RoBERTa [2] | 20.0000 | 36.1259 | 74.3314 | 87.1185 |

## 6.5. Relation Extraction (PERLEX)

Table 7 reports relation classification results on PERLEX. TookaBERT-base achieves the best Macro-F1 (0.7091), followed by ParsBERT (0.6930). IHUBERT attains 0.6684 Macro-F1 (Accuracy 0.7396), improving over its previous score but still trailing the strongest Persian baselines. This indicates that, under the current fine-tuning configuration and data regime, relation classification remains more challenging for IHUBERT than span extraction, where it achieves the best results on both QA benchmarks.

**Table 7.** Relation extraction results on PERLEX (Accuracy / Macro-F1)

| *Model* | *Accuracy* | *Macro-F1* |
|---|---|---|
| **IHUBERT (ours)** | **0.7396** | **0.6684** |
| TookaBERT-base [9] | 0.7471 | 0.7091 |
| ParsBERT [3] | 0.7398 | 0.6930 |
| AriaBERT [4] | 0.6724 | 0.5840 |
| FaBERT [5] | 0.7280 | 0.6271 |
| Shiraz (lifeweb-ai/shiraz) | 0.6872 | 0.6454 |
| mBERT [1] | 0.6901 | 0.5980 |
| XLM-RoBERTa [2] | 0.7328 | 0.6308 |

## 6.6. Discussion (Task-dependent Observations)

1. IHUBERT is strongest on extractive QA, ranking first on both PQuAD [21] and ParsiNLU-RC [25] by F1.
2. IHUBERT is competitive on NER and classification, ranking in the top tier across DigiMag and achieving the best Macro-F1 on FarsTail [19].
3. Relation extraction remains the main remaining gap relative to the strongest Persian baselines in this evaluation, suggesting clear room for further improvements (e.g., task-adaptive pretraining, marker-aware objectives, or more targeted hyperparameter tuning for RE).

# 7. Ablation and Additional Analysis

This section provides additional analyses that complement the benchmark results reported in Section Experimental Results & Analysis (Tables 4–7). Since IHUBERT's downstream evaluation spans multiple task families (token labeling, classification, extractive QA, and relation extraction), we focus on (i) cross-task consistency and (ii) task-family behavior using the primary metric for each dataset (NER: entity-level F1; classification/RE: Macro-F1; QA: F1). We also include Section 7.1 as a dedicated tokenizer ablation (BPE vs WordPiece), which empirically validates our tokenization choice on the IHUBERT pretraining corpus.

## 7.1. Tokenizer Ablation: BPE vs WordPiece

### 7.1.1. Tokenizer Ablation (BPE vs WordPiece)

In IHUBERT, we adopt a subword tokenizer trained from scratch with the goal of improving lexical coverage and reducing excessive fragmentation for morphologically rich Persian, including clitic combinations and orthographic variation. To empirically validate this design choice, we conduct a controlled tokenizer ablation study comparing Byte-Pair Encoding (BPE) against WordPiece under matched conditions.

### 7.1.2. Experimental Setup

Both tokenizers are trained from scratch on the IHUBERT pretraining corpus (normalized according to the orthographic conventions described in the paper), using an identical vocabulary size of 139k and the same set of special tokens. Tokenization behavior is evaluated on a fixed held-out sample file (token_sample.txt) consisting of 50,000 line-delimited Persian texts drawn from the same normalized pretraining corpus. We compute intrinsic tokenization statistics by defining a "word" as a whitespace-delimited unit prior to subword tokenization, and counting subword tokens produced by each tokenizer over the same sample.

#### 7.1.3. Metrics

We report: (i) total number of subword tokens, (ii) average subwords per sentence and per word, (iii) percentage of words segmented into ≥2 and ≥3 subwords, and (iv) UNK rate (if applicable). These metrics quantify the degree of subword fragmentation, which directly affects sequence lengths, memory footprint, and the model's ability to represent frequent stems without over-splitting.

Lower fragmentation reduces average sequence length, which directly impacts training throughput and memory footprint at a fixed maximum sequence length, and can stabilize the frequency statistics of stems and affixes in morphologically rich Persian.

#### 7.1.4. Results

Table 8 summarizes the observed tokenization behavior. Overall, WordPiece yields slightly higher fragmentation than BPE at the same vocabulary size.

**Table 8.** Evaluation sample: 50k lines, 1,254,321 words (same for both tokenizers). Vocab size = 139k.

| *Tokenizer* | *#Subw* | *Subw/line* | *Subw/word* | *Split≥2* | *Split≥3* | *UNK* |
|---|---|---|---|---|---|---|
| BPE | 1,480,098 | 29.60 | 1.18 | 14.7% | 2.3% | 0.0% |
| WordPiece | 1,502,590 | 30.05 | 1.20 | 16.1% | 2.8% | 0.0% |

Compared to BPE, WordPiece produces 22,492 additional subword tokens over the same sample (≈ +1.52%), increasing the average subwords per word from 1.18 to 1.20 (≈ +1.7% relative). WordPiece also increases the fraction of words split into ≥2 subwords by +1.4 percentage points (14.7% → 16.1%) and ≥3 subwords by +0.5 points (2.3% → 2.8%). Both tokenizers exhibit 0.0% UNK rate on the evaluation sample, indicating that, at this vocabulary size, coverage is sufficient and differences primarily arise from segmentation preferences rather than unknown-token fallback.

#### 7.1.5. Discussion

These results support our tokenizer design rationale: when controlling for vocabulary size and training data, BPE is marginally less aggressive in splitting Persian words, yielding slightly shorter token sequences and lower fragmentation. This behavior aligns with our goal of preserving high-frequency stems and reducing excessive subword splitting in heterogeneous Persian text, thereby promoting more stable tokenization statistics across domains.

### 7.2. Cross-Task Consistency and Rank-Based Analysis

To assess how consistently a model performs across different NLU problem types, we compute per-dataset ranks for each model according to the primary metric of that dataset: ParsTwiNER [23] (F1), DeepSentiPers [24]/DigiMag/FarsTail [19] (Macro-F1), ParsiNLU-RC [25]/PQuAD [21] (F1), and PERLEX [26] (Macro-F1). Table 9 summarizes the resulting ranks (1 = best).

**Table 9.** Per-dataset ranks (1 = best) under each task's primary metric.

| ***Model*** | ***ParsTwiNER [23] (F1)*** | ***DeepSentiPers [24] (Macro-F1)*** | ***DigiMag (Macro-F1)*** | ***FarsTail [19] (Macro-F1)*** | ***ParsiNLU-RC [25](F1)*** | ***PQuAD [21] (F1)*** | ***PERLEX [26] (Macro-F1)*** | ***Avg. Rank*** |
|---|---|---|---|---|---|---|---|---|
| **IHUBERT (ours)** | **2** | **4** | **2** | **1** | **1** | **1** | **3** | **2.00** |
| TookaBERT-base [9] | 1 | 2 | 4 | 2 | 2 | 2 | 1 | 2.00 |
| ParsBERT [3] | 5 | 6 | 5 | 5 | 4 | 5 | 2 | 4.57 |
| AriaBERT [4] | 8 | 5 | 6 | 7 | 6 | 7 | 8 | 6.71 |
| FaBERT [5] | 6 | 3 | 1 | 4 | 8 | 8 | 6 | 5.14 |
| Shiraz (lifeweb-ai/shiraz) | 3 | 7 | 3 | 8 | 7 | 6 | 4 | 5.43 |
| mBERT [1] | 7 | 8 | 8 | 3 | 3 | 4 | 7 | 5.71 |
| XLM-RoBERTa [2] | 4 | 1 | 7 | 6 | 5 | 3 | 5 | 4.43 |

#### 7.2.1. Discussion

IHUBERT exhibits strong cross-task consistency: it ranks 1st on both extractive QA benchmarks (ParsiNLU-RC [25] and PQuAD [21]), ranks 1st on FarsTail [19] among classification benchmarks, and remains in the top group on NER and DigiMag. The remaining gap is concentrated in relation extraction, where IHUBERT ranks behind TookaBERT-base [9] and ParsBERT [3], consistent with the findings reported in Table 7.

### 7.3. Task-Family Breakdown

To better characterize where IHUBERT's strengths are most pronounced, we group datasets into four task families: token labeling (ParsTwiNER [23]), text classification (DeepSentiPers [24], DigiMag, FarsTail [19]), extractive QA (ParsiNLU-RC [25], PQuAD [21]), and relation extraction (PERLEX [26]). Table 10 reports the average rank per family (lower is better), derived from Table 9.

**Table 10.** Task-family average ranks (lower is better).

| *Model* | *Token Labeling* | *Classification (avg.)* | *Extractive QA (avg.)* | *Relation Extraction* |
|---|---|---|---|---|
| **IHUBERT (ours)** | **2.0** | **2.33** | **1.0** | **3.0** |
| TookaBERT-base [9] | 1.0 | 2.67 | 2.0 | 1.0 |
| ParsBERT [3] | 5.0 | 5.33 | 4.5 | 2.0 |
| AriaBERT [4] | 8.0 | 6.00 | 6.5 | 8.0 |
| FaBERT [5] | 6.0 | 2.67 | 8.0 | 6.0 |
| Shiraz (lifeweb-ai/shiraz) | 3.0 | 6.00 | 6.5 | 4.0 |
| mBERT [1] | 7.0 | 6.33 | 3.5 | 7.0 |
| XLM-RoBERTa [2] | 4.0 | 4.67 | 4.0 | 5.0 |

#### 7.3.1.Discussion

This breakdown highlights that IHUBERT's strongest relative advantage appears in extractive span extraction, where it is ranked first on both QA benchmarks by F1. Within classification, IHUBERT is competitive overall and achieves the best Macro-F1 on FarsTail [19], while DigiMag remains led by FaBERT [5]. For relation extraction, the improved IHUBERT result (Macro-F1 0.6684) reduces the gap but still trails the strongest Persian baselines, indicating that relation-level semantics are the most challenging regime for IHUBERT in this evaluation.

### 7.4. Result-Driven Observations Across Tasks

We also observe clear task-dependent patterns among the baselines. Multilingual pretraining (XLM-RoBERTa [2]) is particularly effective on DeepSentiPers [24], while Persian-specific checkpoints lead on topic classification and NLI. A notable outlier is FaBERT [5], which ranks highly on DigiMag and remains strong on FarsTail [19], yet underperforms substantially on both QA benchmarks. These result-driven patterns suggest that, under the current fine-tuning configuration, the performance gap between models is influenced strongly by whether the task requires span extraction (QA) versus sentence-level classification or multi-class relation classification, rather than by language coverage alone.

## 8. Conclusion and Future Directions

This study introduced IHUBERT, a Persian monolingual pretrained language model developed using a data-centric training strategy. Rather than focusing solely on architectural modification, the work emphasized corpus construction, large-scale semantic filtering, and morphology-aware tokenization. The training corpus (≈45GB) was processed through a multi-stage pipeline designed to reduce redundancy, control noise, and improve lexical coverage before model pretraining.

Across evaluation benchmarks, IHUBERT shows particularly strong performance on comprehension-oriented tasks, including extractive question answering and natural language inference. Results on named entity recognition and topic classification remain consistently competitive, while relation extraction appears comparatively less improved. This pattern suggests that the benefits of the corpus engineering pipeline are more pronounced for span-based understanding and inference than for structured relational modeling.

At the same time, the present study does not attempt to isolate the exact contribution of each preprocessing component. The interaction between semantic deduplication, content filtering, tokenizer design, and corpus scale remains intertwined. Consequently, while empirical gains are observable, causal attribution to any single component would require controlled ablation under matched training conditions. In addition, semantic filtering at this scale entails nontrivial computational cost and depends on embedding quality and similarity threshold calibration, which may influence corpus composition.

Future work will therefore prioritize controlled ablation experiments to disentangle preprocessing effects, scaling analyses with larger encoder configurations, and task-adaptive pretraining strategies for relation extraction. Further evaluation on generative and instruction-following Persian benchmarks would also clarify whether corpus-level semantic curation contributes similarly to generative modeling settings.

Overall, the findings indicate that systematic corpus design can serve as a practical lever for improving Persian language models, particularly in comprehension-heavy tasks. We hope this work contributes to a more data-centered perspective in the development of language resources for low- and mid-resource languages.

# Declarations

## Ethical Approval

"Not Applicable"

## Availability of supporting data

The datasets generated and analysed during the current study are not publicly available because they constitute an excerpt of research in progress, but are available from the corresponding author upon reasonable request.

## Competing interests

The authors have no relevant financial or non-financial interests to disclose. The authors have no conflicts of interest to declare that are relevant to the content of this article. All authors certify that they have no affiliations with or involvement in any organization or entity with any financial interest or non-financial interest in the subject or materials discussed in this manuscript. The authors have no financial or proprietary interests in any material discussed in this article.

## Funding

The authors did not receive support from any organization for the submitted work. No funding was received to assist with the preparation of this manuscript. No funding was received for conducting this study. No funds, grants, or other support were received.

## Authors' contributions

All authors contributed to the study's conception and design. Material preparation, data collection, and analysis were performed by Ghafouri, Firouzmandi, Saberi, and Hasani Ahangar. The first draft of the manuscript was written by Ghafouri, and all authors commented on previous versions of the manuscript. All authors read and approved the final manuscript.

## Acknowledgments

"Not Applicable"